%% file: egpaper_final.tex
\documentclass[10pt,twocolumn,letterpaper]{article}

\usepackage{cvpr}
\usepackage{times}
\usepackage{epsfig}
\usepackage{graphicx}
\usepackage{amsmath}
\usepackage{amssymb}
\usepackage{subcaption}
\usepackage{booktabs}
\usepackage[symbol]{footmisc}

\cvprfinalcopy 

\def\cvprPaperID{****} 

\begin{document}

\title{2nd Place Solution to the GQA Challenge 2019}
\author{Shijie Geng$^{1}$\thanks{indicates equal contribution} \qquad Ji Zhang$^{1*}$ \qquad Hang Zhang$^2$ \qquad Ahmed Elgammal$^1$ \qquad Dimitris N. Metaxas$^1$\\
$^1$Rutgers University \qquad $^2$Amazon AI\\
{\tt\small \{sg1309,jz462,dnm\}@rutgers.edu} \qquad {\tt\small \{zhanghang0704,elgammal\}@gmail.com} 
}

\maketitle

\begin{abstract}
   We present a simple method that achieves unexpectedly superior performance for Complex Reasoning involved Visual Question Answering. Our solution collects statistical features from high-frequency words of all the questions asked about an image and use them as accurate knowledge for answering further questions of the same image. We are fully aware that this setting is not ubiquitously applicable, and in a more common setting one should assume the questions are asked separately and they cannot be gathered to obtain a knowledge base. Nonetheless, we use this method as an evidence to demonstrate our observation that the bottleneck effect is more severe on the feature extraction part than it is on the knowledge reasoning part. We show significant gaps when using the same reasoning model with 1) ground-truth features; 2) statistical features; 3) detected features from completely learned detectors, and analyze what these gaps mean to researches on visual reasoning topics. Our model with the statistical features achieves the 2nd place in the GQA Challenge 2019.
\end{abstract}

\section{Introduction}

It is well known that Visual Question Answering (VQA) problems is challenging due to the fact that it requires not only accuracy on extracting semantic knowledge from an image such as objects, attributes and relationships, but also a comprehensive ability to conduct reasoning over the knowledge set. Clearly a two-step task as it is, a natural question is which matters more? The answer seems clearer as more recent datasets came out followed by methods that perform reasonably well on them. As one of the earliest, the VQA~\cite{antol2015vqa} and VQA 2.0~\cite{balanced_vqa_v2} datasets directly target QA on natural images, where the latter is more balanced than the former so that questions have less statistical bias that can be leveraged to give direct answers without even looking at the images. The CLEVR dataset~\cite{johnson2017clevr} goes a little further by asking complex logical reasoning questions on image of simple shapes. It is well known as a benchmark to test a model's ability to reason from questions towards answers.

While we have witnessed tremendous success on the CLEVR dataset where recent methods are able to have close-to-perfect performance (98.9\% in~\cite{hudson2018compositional} and 99.8\% in~\cite{yi2018neural} ), results of these methods on natural images are notably lower. What is the main cause, the intrinsic difficulty of perceptual ability on natural images, or the potentially more complex logical processes on these images? This paper provides some observations and insights to this question.

\section{Methods}

\begin{figure}[t!]
 \centering
 \includegraphics[height=5.4cm]{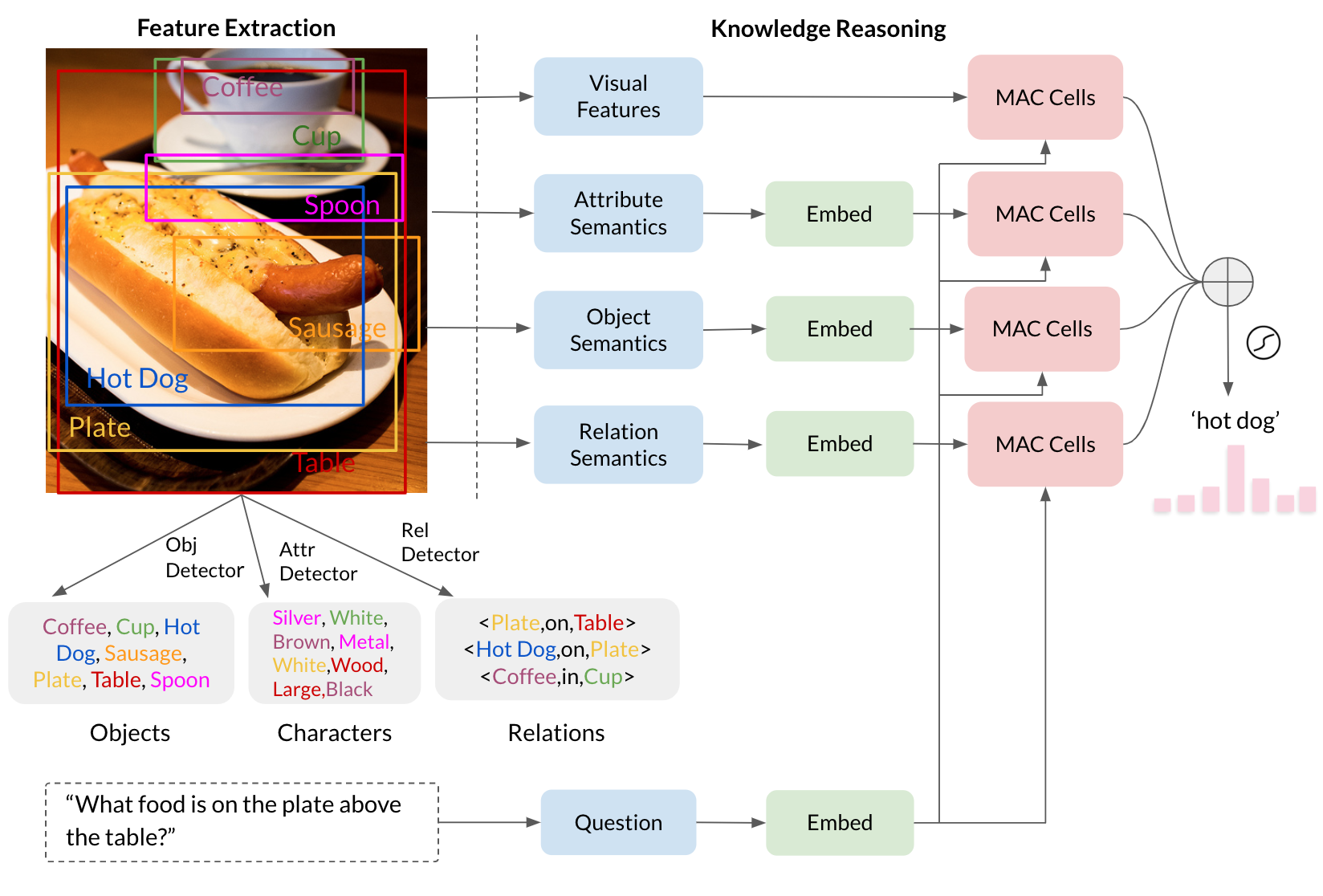}
 \vspace{-5pt}
 \caption{The overview of our model.}
\label{fig:model}
\vspace{-10pt}
\end{figure}

As shown in Figure~\ref{fig:model}, our model is consisted of two modules: feature extraction and knowledge reasoning. The latter takes the output of the former as input, i.e., knowledge of the given image and directly outputs the answer.

For the feature extraction module, we train separate detectors for object detection, attribute detection and relationship detection~\cite{lu2016visual,Zhuang_2017_ICCV,zhang2017visual,dai2017detecting,yu17iccv,Yin_2018_ECCV,LiCVPR2017,xu2017scenegraph,zellers2018neural,zhang2017proposal,zhang2018introduction,zhang2018neurips,zhang2019large,zhang2019graphical}. We use Faster-RCNN~\cite{ren2015faster} for objects, and build a similar system with~\cite{zhang2019graphical} for attributes and relationships. We use several techniques to alleviate the issues caused by large category spaces: 1) We remove plural words in the object categories, which increases mAP@50 from 8.42 to 10.08. The intuition is that turning plurals to singles does not change the semantic meaning but only has fine-grained visual difference, which is in fact of little importance in the GQA dataset; 2) For each output category, if another category is its hypernym, we output it as well since a true category always entails its hypernyms being true. We achieve this by checking the category in the WordNet tree and finding its fore-parent nodes along the upward path to the root. We do this for both object and attribute categories. Note that this is a step only during testing. During training we still treat hypernyms as equally distinguished categories. 3) For attributes, we separate them into two disjoint groups, one for the adjectives and one for the non-adjectives. This is because learning adjectives and non-adjectives requires a model to focus on different property of objects, e.g., adjectives usually describe colors, textures or sizes, while non-adjectives are often about materials or components. This separation is done by checking whether a word has non-empty adjective synsets. 4) For the predicates of relationships, we separate them into three groups: spatial, interaction and others. Interaction predicates can be filtered out by checking whether a word has non-empty verb synsets, while the other two groups are selected manually. Once the objects, attributes and relationships are detected, their labels are embedded into 300D vectors by GloVe~\cite{pennington2014glove}. Including object CNN features provide by the GQA dataset, we have totally four types of features that provide various aspects of the knowledge about an image. 
For the visual reasoning module, we use Compositional Attention Networks for Machine Reasoning (MAC)~\cite{hudson2018compositional} as the backbone of our reasoning module. We use a late fusion strategy that feeds each of the four features into one independent branch with the MAC structure, then add up the output logits of the four branches followed by a Sigmoid on the sum to obtain a probability distribution over the dictionary of all answers, which is a common strategy adopted by many VQA systems. 

\section{Experiments}
\noindent \textbf{Ablation Analysis on the Feature Type.} Table \ref{tab:val} presents results using our detected features, statistical features and ground truth features with two reasoning models: MAC~\cite{hudson2018compositional} and multi-stream cross attentional model (MS-CA) from TVQA~\cite{lei2018tvqa}. The detected features are extracted by our trained object, attribute and relationship detectors, while the statistical features are obtained by first counting the frequency of each word of all the questions about one image, then taking only those words with frequencies higher than a threshold (which is set as 10 in this paper). The MS-CA model is a strong baseline provided by~\cite{lei2018tvqa} for video question answering, and we modify it for our visual reasoning based question answering. The purpose of this ablation is to compare the gaps between different features when the model is fixed, and compare the gaps between different models when the features are fixed. We can see that statistical features lead to results that are close to those from ground-truth features, which we believe is due to the fact that some questions can provide useful information for other questions, i.e., there are valuable facts hidden in many questions. The gap between detected features and statistical features are obviously larger than the one between statistical features and ground-truth features, indicating the intrinsic difficulty of learning to accurately recognize useful visual knowledge from natural images. Another comparison can be done between the two different models when features are fixed. The gaps between the two models for each of the three features are relatively smaller than the gaps between different features when the model is fixed, suggesting that the difference of reasoning models might not be as important as difference of features. This acts as a support to our claim that the bottleneck for success of visual question answering is in fact more on the inaccuracy of feature extraction than on the lack of reasoning ability.

\input{tab/feature_val.tex}

{\small
\bibliographystyle{ieee}
\bibliography{egbib}
}

\end{document}

%% file: tab/feature_val.tex
\begin{table}[t]
\centering
\begin{tabular}{lcc}
\toprule
\noalign{\smallskip}
Features & Val Acc. & Test Acc.\\
\noalign{\smallskip}
\hline
\noalign{\smallskip}
Detected (MS-CA) & 55.48 & 49.74\\
Detected (MAC) & 62.98 & 56.21\\
Stats (MS-CA) & 69.67 & 64.48\\
Stats (MAC) & 76.15 & 70.23\\
Groundtruth (MS-CA) & 79.20 & -\\
Groundtruth (MAC) & 81.14 & -\\
\noalign{\smallskip}
\bottomrule
\end{tabular}
\caption{Ablation study for using different features as the knowledge base on validation split. }
\label{tab:val}
\vspace{-12pt}
\end{table}